\documentclass[a4paper]{article}

\usepackage{geometry}
\usepackage{graphicx, hyperref, setspace, amsmath, amssymb, titlesec, fancyhdr, multicol, parskip, indentfirst, etoolbox, caption, cite, xcolor}
\usepackage{algorithm}
\usepackage{algpseudocode}
\usepackage{cuted}
\usepackage{booktabs}
\usepackage{float}

\titleformat{\section}{\centering\large\scshape}{\thesection}{1em}{}
\titleformat{\subsection}{\normalsize\bfseries}{\thesubsection.}{1em}{}
\titlespacing{\section}{0pt}{6pt}{6pt}
\titlespacing{\subsection}{0pt}{6pt}{6pt}
\titlespacing{\subsubsection}{0pt}{6pt}{6pt}

\title{Real-Time Localization Framework for Autonomous Basketball Robots}

\date{} 
\renewcommand{\thesection}{\Roman{section}.}
\renewcommand{\thesubsection}{\textit{\Alph{subsection}.}}
\renewcommand{\thesubsubsection}{\textit{\arabic{subsubsection}.}}

\titleformat{\subsection}{\normalfont\large\itshape}{\thesubsection}{1em}{}
\titleformat{\subsubsection}{\normalfont\itshape}{\thesubsubsection}{1em}{}

\makeatletter
\renewcommand{\maketitle}{{
  \begin{center}
    \noindent\rule{\textwidth}{1pt}\par
    \vspace{0.3em}
    {\LARGE \bfseries \@title \par}
    \vspace{0.3em}
    \noindent\rule{\textwidth}{1pt}\par
  \end{center}
}}
\makeatother

\begin{document}

\maketitle
\vspace{1cm}

\begin{multicols}{2}
    \centering
    \textbf{Naren Medarametla}\\
    \textit{School of Computer Science Engineering}\\
    \textit{Vellore Institute of Technology}\\
    \textit{Chennai, India}\\
    \texttt{naren.medarametla2023@vitstudent.ac.in}
    \vfill\null

    \columnbreak

    \textbf{Sreejon Mondal}\\
    \textit{School of Electrical and Electronics Engineering}\\
    \textit{Vellore Institute of Technology}\\
    \textit{Chennai, India}\\
    \texttt{sreejon.mondal2023@vitstudent.ac.in}
    \vfill\null
\end{multicols}

\singlespacing
\setlength{\parskip}{6pt}
\setlength{\parindent}{0.5cm}

\begin{multicols}{2}
% \raggedcolumns
\setlength{\columnsep}{0.5cm}

\noindent \textbf{\textit{Abstract---}
Localization is a fundamental capability for autonomous robots, enabling them 
to operate effectively in dynamic environments.
In Robocon 2025, accurate and 
reliable localization is crucial for improving shooting precision, avoiding 
collisions with other robots, and navigating the competition field efficiently.
In this paper, we propose a hybrid localization algorithm that integrates classical 
techniques with learning based methods that rely solely on visual data from the court’s
floor to achieve self-localization on the basketball field.
}

\small  
\noindent \textbf{
  \textit{Keywords---}\textit{Robot Localization, Autonomous Navigation, Neural Networks, Robocon}}

\section{Introduction}
\par \noindent
The DD Robocon is the Indian national entry level competition to Asia Pacific Robot Contest (ABU Robocon).
The objective of the competition is for college teams to design, build, and operate robots to complete certain tasks.
The tasks vary each year and are usually based on cultural, historical, or technological themes related to the host country of the international contest.
For Robocon 2025, the theme was Robot Basketball, where two robots had to play the game of basketball following specific rules.
The arena consisted of a court of dimensions 15m $\times$ 8m and baskets at a height of 2.4m.
Due to the nature of the game, the robots had to move quickly and unpredictably, and take shots from various positions on the court.
The robots had to have robust and real time self localization determine it's distance from the basket.
In this paper we propose a vision based localization system based on the white field lines of the basketball court.
\par \noindent
\textbf{Section II} reviews existing methods and prior research related to this work.
\textbf{Section III} provides a detailed description of our proposed algorithm, approach, and model architecture.
\textbf{Section IV} provides the results obtained from our experiments. 
\textbf{Section V} discusses previous approaches which failed.
\textbf{Section VI} evaluates the accuracy of our approach and discusses potential directions for future work.
\section{Related Work}
\par \noindent In vision based self localization for dynamic field sports, 
researchers have explored complementary lines of work that balance geometric modeling, 
probabilistic estimation, and learned perception.
Liu et al.~\cite{liu2009} introduced a
pose point matching method for omnidirectional cameras in which white field lines are 
detected from brightness changes, corrected for lens effects, and matched against a 
field model to recover position and heading in real time with errors reported on the 
order of tens of centimeters.
Ribeiro et al.~\cite{ribeiro2016} accelerated global 
relocalization by precomputing a distance map of the pitch and then searching 
the error surface between observed line points and this map under a known heading, 
which makes frequent global updates practical when calibration and line extraction
are reliable.
Watanabe et al.~\cite{watanabe2020} framed model matching as a search 
problem and optimized the robot pose with a genetic algorithm driven by omnidirectional 
white line features, which helps in scenes with sparse or ambiguous cues while requiring 
careful parameterization to remain real time.
\par \noindent Lu et al.~\cite{lu2013} reported a robust 
real time pipeline based on omnidirectional vision that emphasizes efficient feature 
extraction and stable operation during matches, though performance naturally depends 
on lighting and occlusions typical of indoor arenas.
On humanoid platforms, Tian et 
al.~\cite{tian2010} adapted Monte Carlo Localization to fisheye optics and bipedal 
gait by designing a motion model for oscillatory head movement and a vision model 
for field landmarks, together with a resampling strategy that stabilized estimates 
on limited hardware.
Fusion based designs combine complementary sensing to limit drift 
and maintain responsiveness, Ismail et al.~\cite{ismail2012} fused encoder and gyro 
odometry with omnivision in a particle filter so that fast motion updates are anchored
by drift free visual corrections even on symmetric fields.
\par \noindent More recent work uses inertial
constraints to stabilize vision before geometric reasoning, Nadiri et al.~\cite{nadiri2025}
filtered IMU orientation and applied inverse perspective mapping to obtain a bird’s eye view
from which field markers provide consistent distances on approximately planar surfaces.
Learning based perception is widely used to localize robots relative to other agents
rather than only to the field, Luo et al.~\cite{luo2020} detected robots with a convolutional
detector and estimated 3D positions through RGBD registration on embedded hardware, improving 
robustness to appearance variation at the cost of additional calibration and power.
\end{multicols} 

\begin{center}
\captionof{table}{Comparison of Related Works}
\label{tab:localization_comparison}
\resizebox{\textwidth}{!}{%
\begin{tabular}{@{}llp{4.5cm}p{4cm}p{4cm}@{}}
\toprule
\textbf{Author \& Year} & \textbf{Camera Type} & \textbf{Methodology} & \textbf{Performance} & \textbf{Limitations} \\ 
\midrule
\addlinespace
\textbf{Liu et al, 2009} & Omnidirectional & Extracts white line feature points from brightness changes and matches them to model “pose points” for position \& heading estimation.
& Achieves approx. 20 fps with approx. 20 cm mean error, robust to changes in goal color.
& Relies on good visibility of white lines, requires careful distortion correction.
\\ 
\addlinespace
\textbf{Ribeiro et al, 2016} & Omnidirectional & Uses precomputed distance maps of field lines, matches observed line points to the least error grid cell given a known heading.
& Allows for very fast global localization and frequent updates.
& Needs an accurate heading and precise field calibration, performance degrades with poor line detection.
\\ 
\addlinespace
\textbf{Watanabe et al, 2020} & Omnidirectional & Combines white line detection with a Genetic Algorithm (GA) to optimize the pose model match.
& Provides robust performance even with sparse line cues and enables an effective global search.
& Requires tuning the GA for real-time performance, has a higher computational cost than analytic matchers.
\\ 
\addlinespace
\textbf{Luo et al, 2019/2020} & RGB + Kinect v2 Depth & A two stage process where YOLOv2 detects robots in RGB, and then Kinect depth data is used to recover their 3D positions.
& High Mean Average Precision (mAP), robust against changes in jersey color.
& Requires an active depth sensor, cross sensor calibration, and has extra power demands.
\\ 
\addlinespace
\textbf{Lu et al, 2013} & Omnidirectional & Employs a robust omnidirectional vision pipeline for feature extraction and localization.
& Has demonstrated real-time robustness in practical applications. & Sensitive to lighting conditions and occlusion.
\\ 
\addlinespace
\textbf{Tian et al, 2010} & Fisheye Lens & Implements a particle filter (Monte Carlo Localization) with vision and motion models, handling distortion and motion oscillation.
& Manages non Gaussian noise and multiple hypotheses, has been proven on hardware.
& Incurs a higher computational load, can suffer from pose aliasing in symmetric fields.
\\ 
\addlinespace
\textbf{Ismail et al, 2012} & Omnidirectional + Odometry & Fuses data from odometry (which is fast but drifts) and omnidirectional vision with MCL (which is drift free but slower).
& Delivers accurate and responsive localization with low latency (approx. 1.6 ms).
& Requires careful time synchronization, MCL remains compute heavy with sparse data.
\\ 
\addlinespace
\textbf{Nadiri et al, 2025} & Omnidirectional + IMU & Uses an IMU stabilized camera with a Bird’s Eye View (BEV) transformation for marker detection.
& Reduces mean error by approximately 9 cm, the BEV simplifies distance measurement.
& Demands reliable IMU calibration, sensitive to errors in camera setup.
\\ 
\addlinespace
\bottomrule
\end{tabular}%
}
\end{center}

\vspace{0.5cm}
\par \noindent
Unlike most of the reviewed approaches, which rely heavily on omnidirectional or fisheye optics to maximize field of view for white line detection, 
our method uses a regular monocular camera while still achieving reliable self localization.
This choice reduces the need for specialized distortion 
correction pipelines, and makes the approach more adaptable to standard camera modules already used in many robotics platforms.
\begin{multicols}{2} % Resume columns after table

\section{Methodology}
\par \noindent Our approach is a two step process that begins with \textbf{Preprocessing} the image,
followed by passing it to the model for \textbf{Inference}.
\subsection{Preprocessing}
\par \noindent The input image having dimensions (640 $\times$ 480 $\times$ 3) is converted from the RGB color space to the HSV color space, then the white regions are masked out using two predefined HSV ranges.
The image is downsampled through a radial scan, flattened, and finally passed through the neural network.
Figure~\ref{fig:flowchart} shows the preprocessing pipline and Algorithm 1 gives the implementation of the downsampling algorithm.
\end{multicols}

\begin{center}
\includegraphics[width=0.8\textwidth]{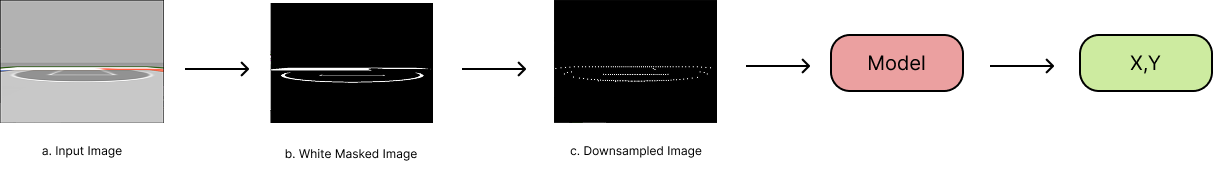}
\captionof{figure}{Preprocessing pipeline}
\label{fig:flowchart}
\end{center}

\begin{multicols}{2}

\begin{algorithm}[H]
  \caption{Downsampling}
\begin{algorithmic}[1]
\Statex \textbf{Input: }Image
\State $H \gets$ Image.height
\State $W \gets$ Image.width
\State $R \gets$ Black Image
\For{angle $\gets$ 0 to 180 step 2}
    \State $lastPixel \gets 0$
    \State $cx \gets \text{W / 2}$
    \State $cy \gets \text{H}$
    \For{d $\gets$ 0 to max(H, W)}
        \State $x \gets cx + d \times \cos(angle)$
        \State $y \gets cy - d \times \sin(angle)$
        \If{$0 \leq x < \text{W} \ \textbf{and} \ 0 \leq y < \text{H}$}
          \State $pixel \gets Image[y][x]$
          \If{lastPixel = 255 and pixel $\neq$ 255}
            \State $R[y][x] \gets 255 $ 
          \EndIf
          \State $lastPixel \gets pixel$
        \EndIf
    \EndFor
\EndFor
\State \textbf{Return} R
\end{algorithmic}
\end{algorithm}

\subsection{Model Architecture}
\par \noindent
The proposed model is a feedforward neural network consisting of a flattening layer followed 
by four fully connected layers. The first 200 pixels from the top of the image are removed 
to reduce the input size, as they do not carry useful information and the pixel values are scaled
down to the range [0, 1].
\par \noindent
The resulting image with dimensions (640 $\times$ 280 $\times$ 1) is flattened into a vector 
of size 1,79,200 and passed through the first linear layer, followed by a ReLU activation function.
The subsequent layers have sizes 1024, 256, and 64, each followed by a ReLU activation.
The final layer outputs a 2 dimensional vector representing the predicted $x$ and $y$ positions of the robot.

\par \noindent
The rationale for using this relatively simple model lies in the simplicity of the input images 
and to reducing inference time.

{ \centering
 \includegraphics[width=0.2\columnwidth]{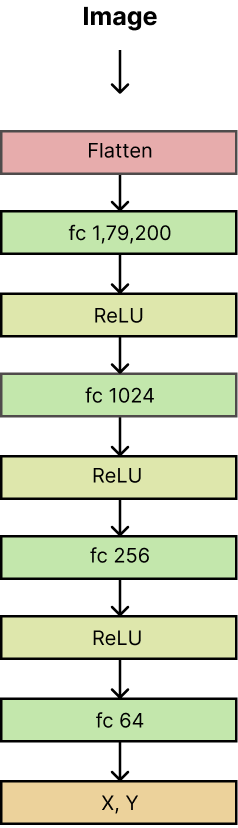}\\
 \captionof{figure}{Flow diagram}\label{fig:model}
}

% \vspace{-0.3cm}
{ \centering
 \includegraphics[width=0.9\columnwidth]{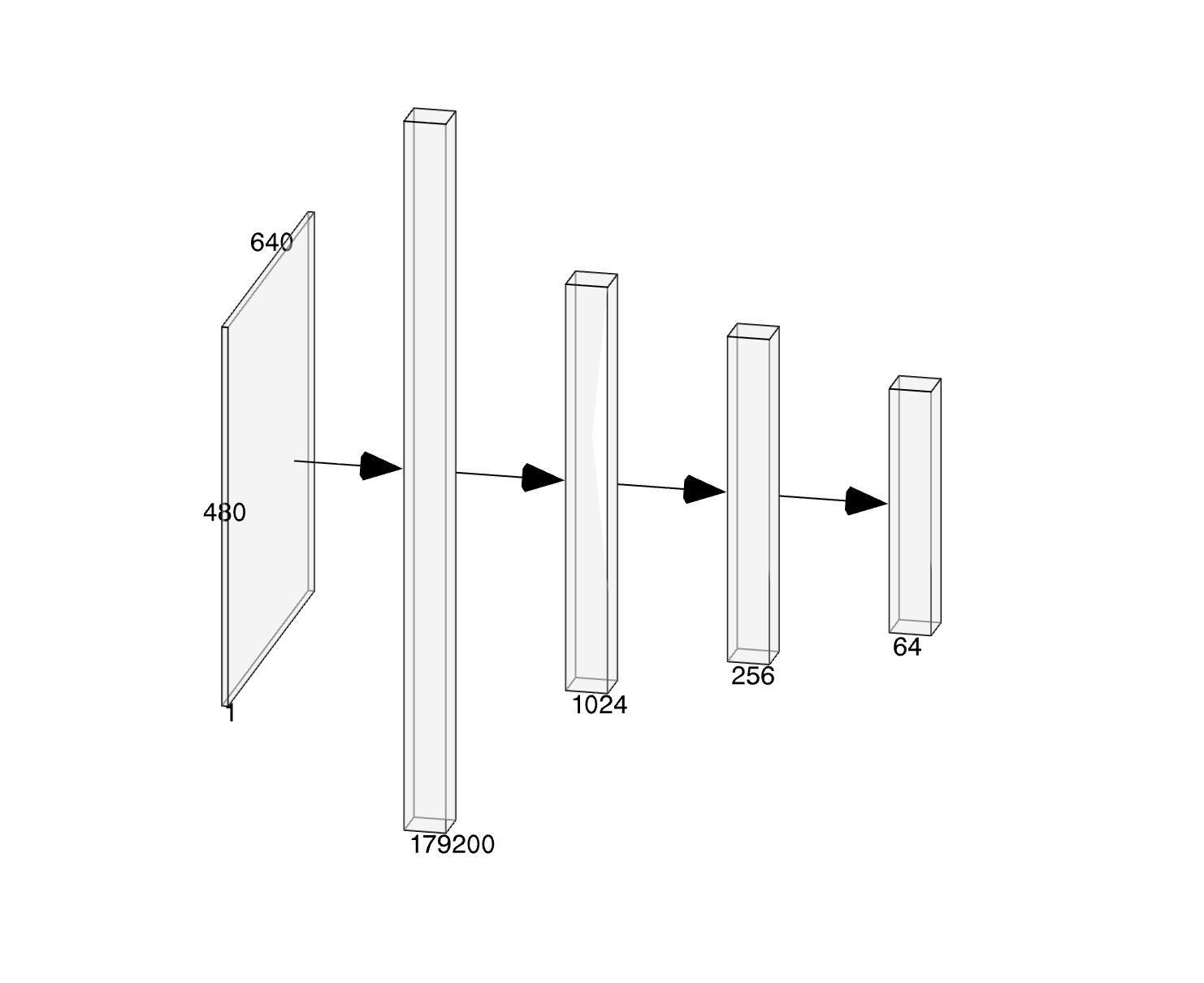}\\
 \captionof{figure}{Architecture diagram}\label{fig:model}
}

\subsection{Dataset} 

\par \noindent
A digital twin of the robot was created and simulated in a replica of the Robocon 2025 arena 
using the Gazebo \cite{gazebo} simulator with the help of ROS2 \cite{macenski2022ros2}.
\par \noindent
The robot was driven through the arena capturing images and corresponding $x$, $y$ coordinates.
A 
TimeSynchronizer was used to synchronize the frame header and the position header.
A total of
6283 images were captured and split into training and test datasets with a 0.9 to 0.1 ratio.
Care was taken to include every part of the arena for an unbiased dataset.
\subsection{Simulation}

\par \noindent The simulation environment was designed to closely replicate the physical conditions of the target field.
A detailed world map of the area was created to serve as the operating environment, and a complete model of the robot was developed using the Unified Robot Description Format (URDF)\cite{tola2024understanding}.
\par\noindent
The simulation was implemented in Gazebo, where the robot's onboard camera was modeled as a virtual sensor using Gazebo camera plugins to replicate image capture.
The robot was teleoperated within the environment using a joystick, leveraging the \texttt{joy} package available in ROS~2 Humble, which enabled real-time manual control and data collection across different positions and orientations in the simulated space.
The images captured in Gazebo were sent back to ROS2 where the model would process them and predict the coordinates.
Figure IV illustrates the block diagram 

{ \centering
 \includegraphics[scale=0.2]{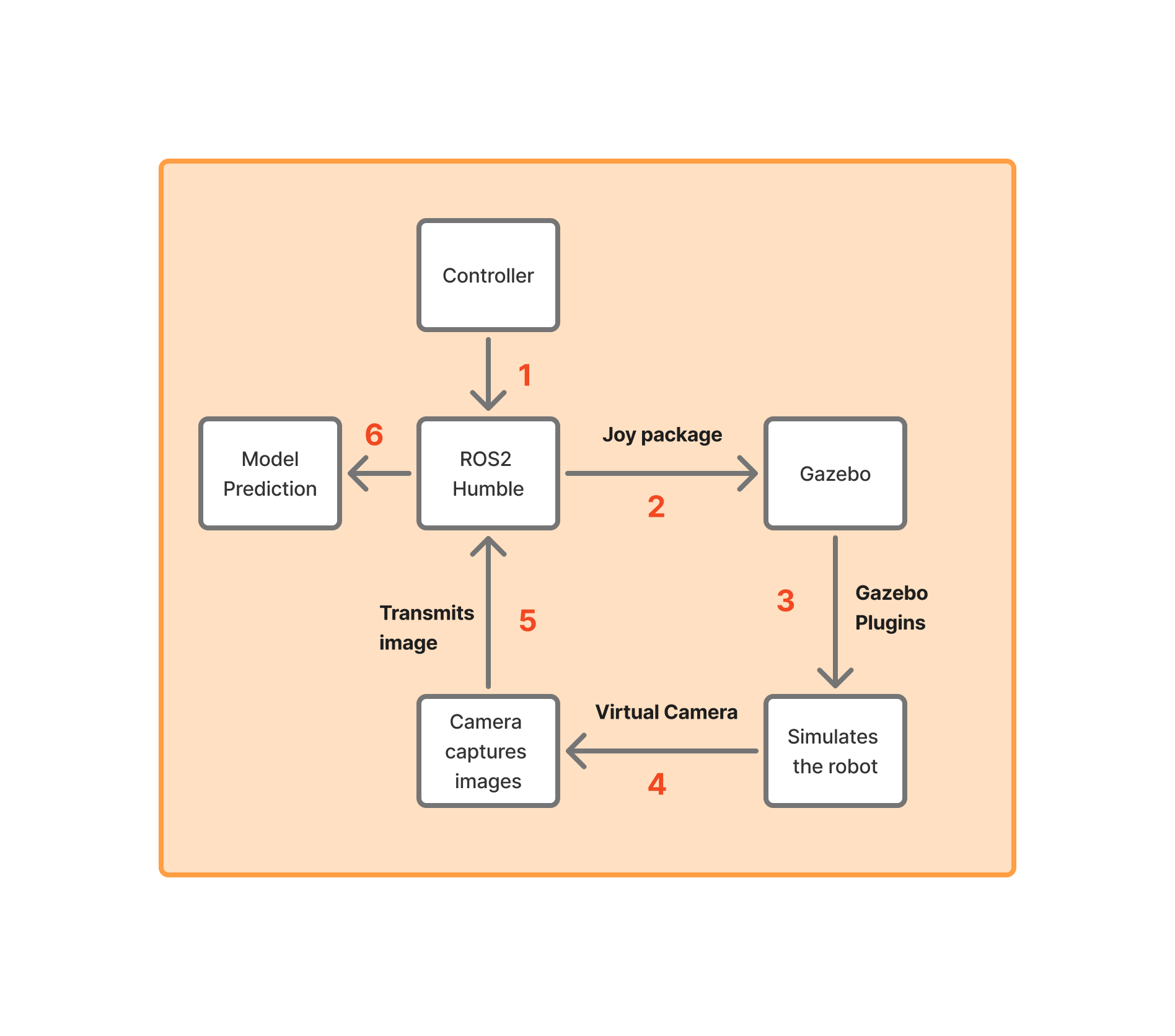}\\
 \captionof{figure}{Blcok Diagram}\label{fig:gazebo}
}

{ \centering
 \includegraphics[scale=0.3]{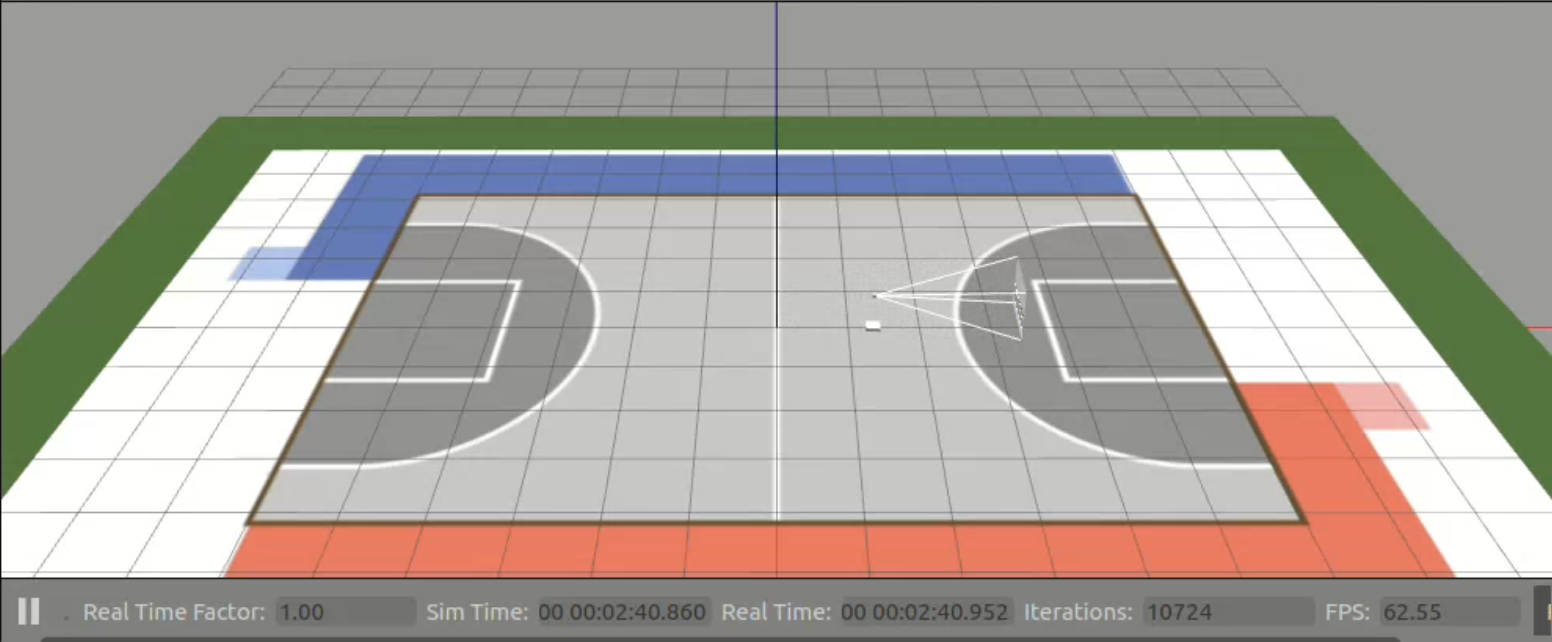}\\
 \captionof{figure}{Gazebo simulation}\label{fig:gazebo}
}

\subsection{Training}
\par \noindent
The model was trained for 15 epochs using an Adam optimizer \cite{kingma2014adam} with an initial learning rate of $10^{-4}$
and a Mean Squared Error (MSE) cost function in batches of size 8. 

\section{Results}
% add repo link as well
\par \noindent
Figure~\ref{fig:loss_curve} depicts the loss at each epoch throughout the training process.
8 independent images at different points of the court were again captured 
from the simulation and fed into the model.
Figure~\ref{fig:comparison} shows the plot between
the ground truth and the prediction made by the model for these 8 images.
\par \noindent
Table~\ref{tab:prediction_loss} illustrates the x and y losses for a corresponding number of test iterations.
Figure~\ref{fig:x_loss} and Figure~\ref{fig:y_loss} show the%  graphical repr% esentation of this loss.
\par \noindent
A Pearson correlation test was performed to determine if the prediction loss was dependent on the number of test iterations. For the x loss, the test yielded a p-value of 0.40. Since this value is well above the significance level of 0.05, we conclude that there is no statistically significant evidence of a relationship, indicating that the x loss is independent of the number of iterations. In contrast, the test for the y loss yielded a p-value of 0.01. This value is below 0.05, indicating a statistically significant negative correlation where the loss tends to decrease as iterations increase. A detailed breakdown of these calculations is provided below.
\par \noindent 
The code for the paper can be found in the following repository:
\href{https://github.com/NarenTheNumpkin/Basketball-robot-localization}{https://github.com/NarenTheNumpkin/Basketball-robot-localization}

\begin{table}[H]
\centering
\captionof{table}{Prediction Loss}\label{tab:prediction_loss}
\begin{tabular}{|c|c|c|c|}
\hline
\textbf{S.No} & \textbf{Iterations} & \textbf{x loss (m)} & \textbf{y loss (m)} \\ \hline
1             & 15                  & 0.1375              & 0.1950                \\ \hline
2             & 25                  & 0.1639              & 0.1861                \\ \hline
3             & 35                  & 0.1498              & 0.1717                \\ \hline
4             & 100                 & 0.1611              & 0.1691                \\ \hline
5             & 200                 & 0.1464              & 0.1519                \\ \hline
6             & 300                 & 0.1412              & 0.1579                \\ \hline
7             & 400                 & 0.1394              & 0.1605                \\ \hline
8             & 500                 & 0.1376              & 0.1588                \\ \hline
9             & 600                 & 0.1401              & 0.1628                \\ \hline
              &                     & \textbf{0.1463}              & \textbf{0.1506}                \\ \hline
\end{tabular}
\end{table}

{ \centering
 \includegraphics[scale=0.2]{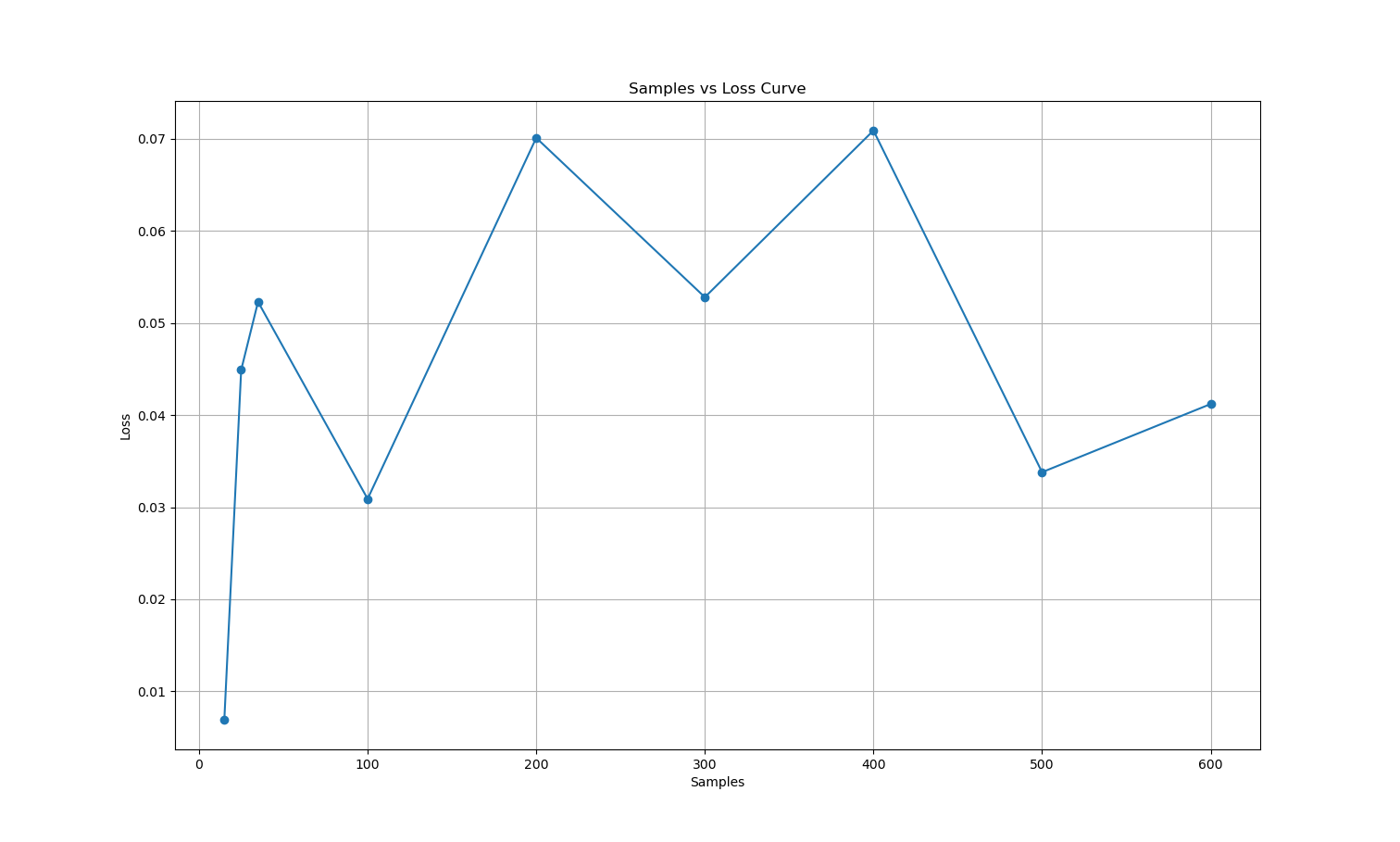}\\
 \captionof{figure}{x loss vs iterations}\label{fig:x_loss}
}

{ \centering
 \includegraphics[scale=0.2]{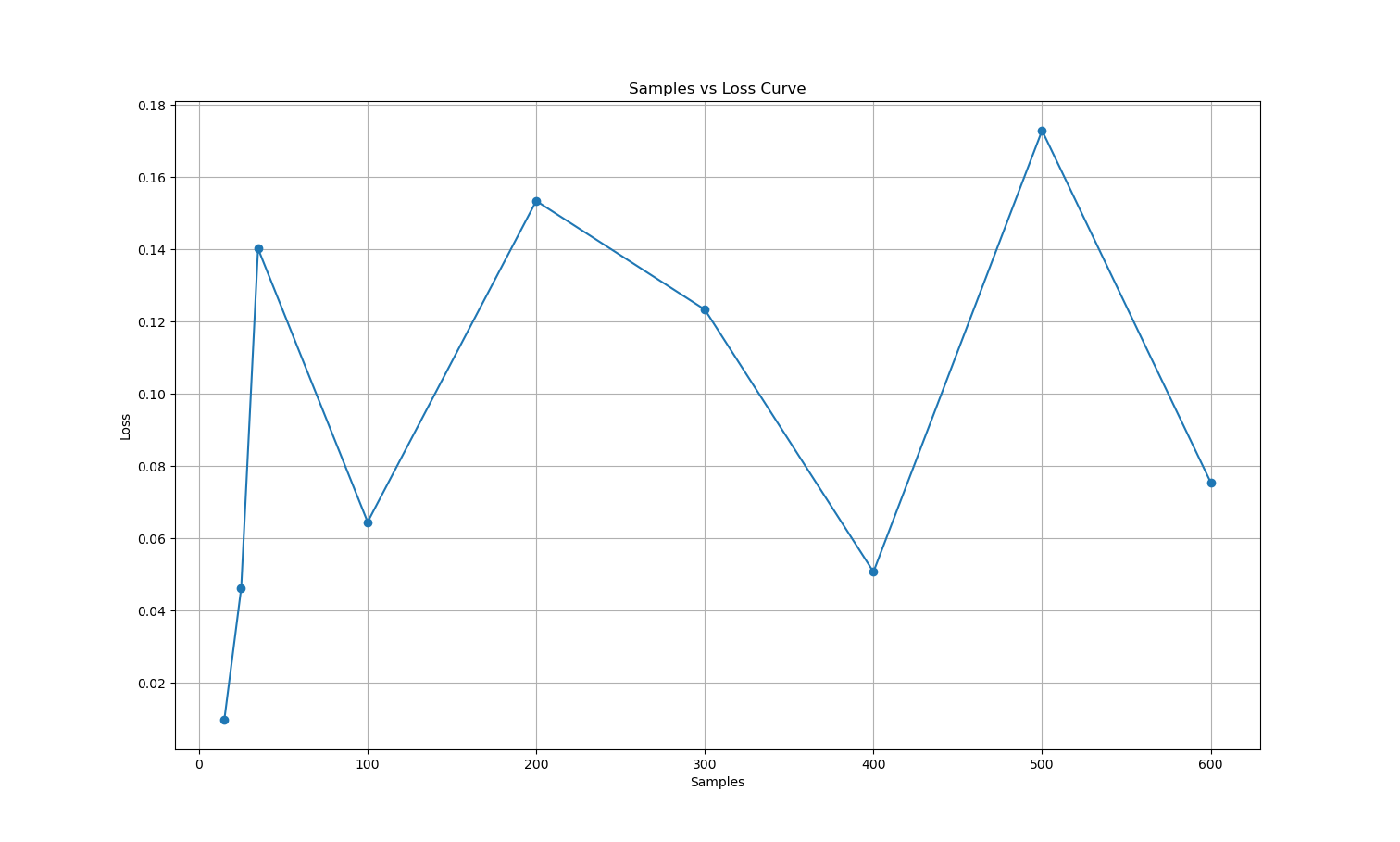}\\
 \captionof{figure}{y loss vs iterations}\label{fig:y_loss}
}

{ \centering
 \includegraphics[scale=0.3]{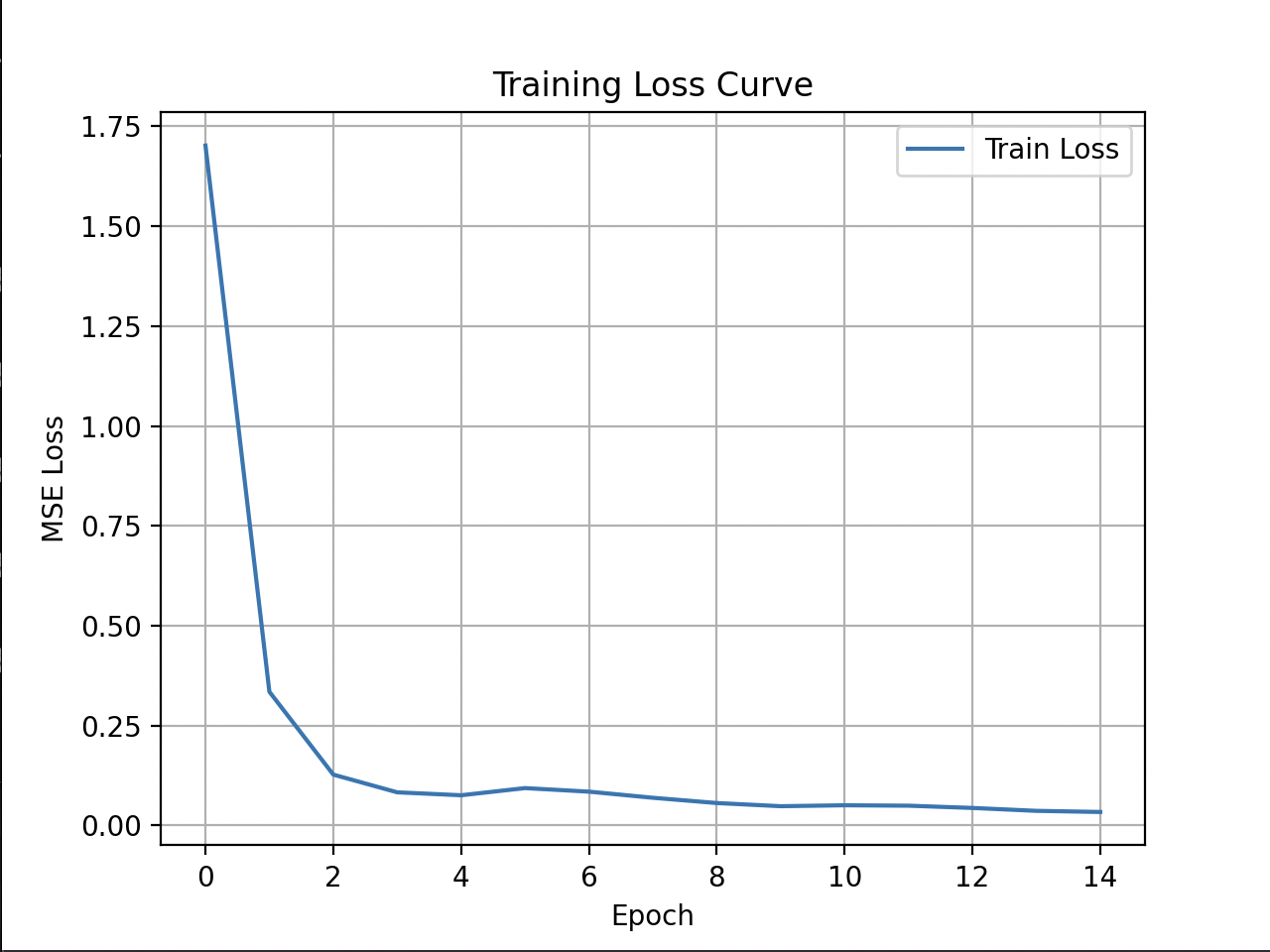}\\
 \captionof{figure}{Loss curve}\label{fig:loss_curve}
}

{ \centering
 \includegraphics[scale=0.18]{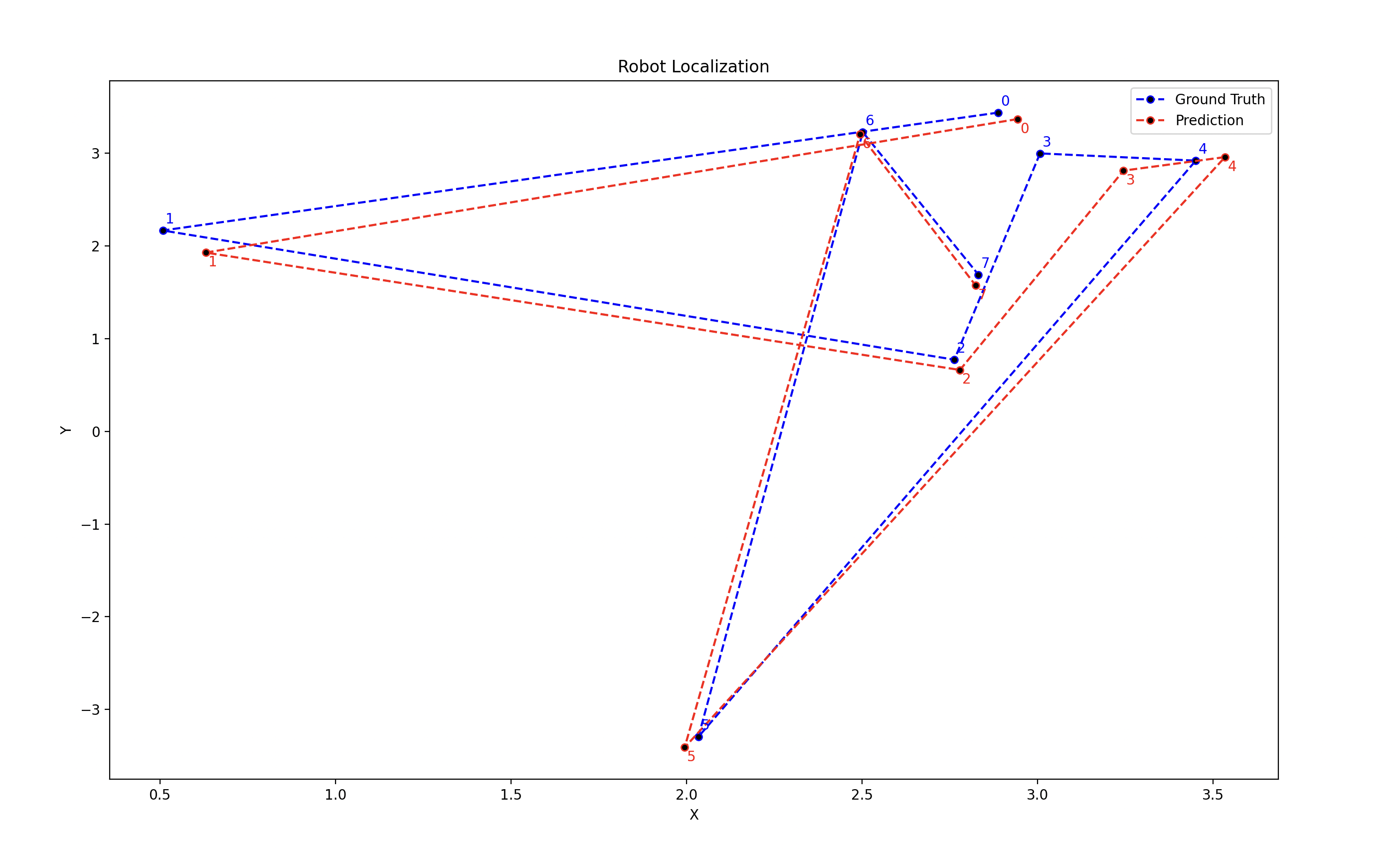}\\
 \captionof{figure}{Ground truth vs prediction}\label{fig:comparison}
}

\subsubsection{Iterations vs. x loss}
\par \noindent
For this test, we used 9 data pairs, obtained a Pearson correlation of $r=-0.32$, and computed the degrees of freedom as $df = 7$ ($n-2$).
The t-statistic is calculated as
$$ t = -0.32 \frac{\sqrt{9-2}}{\sqrt{1 - (-0.32)^2}} = \frac{-0.8467}{0.9474} \approx -0.89 $$
The two tailed critical t-value for $\alpha=0.05$ and df=7 is $\pm2.365$. Since our calculated t-statistic of -0.89 is greater than -2.365, we fail to reject the null hypothesis. The corresponding p value is 0.40. This confirms there is no statistically significant relationship. 

\noindent At $\alpha=0.05$ there is insufficient evidence of a linear association between iterations and x loss. The observed correlation (r = -0.32) is small and likely due to sampling variability rather than a true effect.
\\
\subsubsection{Iterations vs. y loss}
\par \noindent
A corresponding test for iterations vs. y loss yielded $p=0.01<0.05$. Therefore, we reject $H_0$ and conclude there is a statistically significant negative linear relationship between iterations and y loss, indicating that y loss tends to decrease as the number of iterations increases.
$$ t = -0.67 \frac{\sqrt{9-2}}{\sqrt{1 - (-0.67)^2}} = \frac{-1.7727}{0.7424} \approx -2.39 $$

\subsection{Comparision and Analysis}
\par \noindent
Two models were chosen as a baseline, EfficientNet-B\cite{tan2019efficientnet} and MobileNetV2\cite{sandler2018mobilenetv2} due to their ability to run on 
efficiently edge devices which is particularly suitable for faster inference time.
\subsubsection{MobilenetV2}
\par \noindent
MobilenetV2 is built on depthwise separable convolutions and inverted residual blocks with linear bottlenecks. A standard $k{\times}k$ convolution with $M$ input and $N$ output channels over a $D_F{\times}D_F$ map costs
\begin{equation}
\mathrm{Cost}_{\text{std}} = D_K^2 \cdot M \cdot N \cdot D_F^2
\end{equation}
Depthwise separable convolutions reduce this to
\begin{equation}
\mathrm{Cost}_{\text{dw-sep}} = D_K^2 \cdot M \cdot D_F^2 \;+\; M \cdot N \cdot D_F^2
\end{equation}
% With width multiplier $\alpha\!\in\!(0,1]$ and resolution multiplier $\rho\!\in\!(0,1]$ (as in MobilenetV2), the cost is:
% \begin{equation}
% \mathrm{Cost}_{\alpha,\rho} = D_K^2 \cdot (\alpha M) \cdot (\rho D_F)^2 \;+\; (\alpha M)\cdot(\alpha N)\cdot(\rho D_F)^2.
% \end{equation}
A depthwise separable block can be written as
\begin{equation}
y = \mathrm{Conv}_{1\times1}\!\Big(\sigma\!\big(\mathrm{BN}(\mathrm{Conv}_{\text{dw}}(x))\big)\Big)
\end{equation}
where BN is batch normalization and $\sigma$ is a non linear operator,
MobilenetV2 with inverted residual blocks and linear bottleneck has the operator flow
\begin{align}
z &= \phi\!\big(\mathrm{BN}(\mathrm{Conv}^{\text{expand}}_{1\times1}(x))\big) \\
u &= \phi\!\big(\mathrm{BN}(\mathrm{Conv}^{\text{dw}}_{k\times k}(z))\big) \\
y &= \mathrm{BN}(\mathrm{Conv}^{\text{proj}}_{1\times1}(u))
\end{align}

\subsubsection{EfficientNet-B0}
\par \noindent
EfficientNet is a family of lightweight CNN backbones that pair a simple convolutional stem with stacked MBConv stages and a compact head. The network downsamples spatial resolution via staged strides across MBConv blocks and applies squeeze and excitation to recalibrate channels, and adopts SiLU/Swish activations for stable training. A global average pooling layer and a small fully connected head finalize the architecture. Model variants use a compound policy to jointly scale depth, width, and input resolution to meet different latency/accuracy targets without changing the block design.
\par \noindent
The backbone uses MBConv (inverted residual) blocks with squeeze-and-excitation (SE) and SiLU/Swish.
A simplified MBConv with SE
\begin{align}
z &= \phi\!\big(\mathrm{BN}(\mathrm{Conv}^{\text{expand}}_{1\times1}(x))\big), \\
u &= \phi\!\big(\mathrm{BN}(\mathrm{Conv}^{\text{dw}}_{k\times k, s}(z))\big), \\
s &= \sigma\!\big(W_2\,\delta(W_1\,\mathrm{GAP}(u))\big) \odot u, \\
y &= \mathrm{BN}(\mathrm{Conv}^{\text{proj}}_{1\times1}(s)).
\end{align}
Skip connection:
\begin{equation}
\hat y =
\begin{cases}
x + y, & \text{if } s=1 \text{ and } C_{\text{out}}=C_{\text{in}},\\
y, & \text{otherwise.}
\end{cases}
\end{equation}

\begin{table}[H]
\caption{Model Comparison}
\label{tab:model_comparison}
\centering
\small
\setlength{\tabcolsep}{4pt} % tighter columns
\resizebox{\columnwidth}{!}{%
\begin{tabular}{@{}lccc@{}}
\toprule
Metric & \textbf{Feedforward} & \textbf{MobileNetV2} & \textbf{EfficientNet-B0} \\
\midrule
\textbf{MSE} & 0.040 & 3.380 & 3.778 \\
\textbf{Time (ms)}  & 3.283 & 4.213 & 4.373 \\
\textbf{RAM} & 1.3 GB & 387 MB & 323 MB \\
\textbf{Temperature} & 43.62°C & 43.69°C & 43.97°C  \\
\bottomrule
\end{tabular}%
}
\end{table}

\par \noindent
The above table shows the performance of the models on physical hardware on 100 samples. Our feedforward model achieves the lowest MSE and the fastest inference, primarily because its computation mainly involves a few dense matrix multiplications on a flattened preprocessed input. However, this design incurs higher memory usage, flattening a large spatial map into a 179{,}200-dimensional vector and feeding it into wide fully connected layers creates large weight matrices and activation tensors. In contrast, MobileNetV2 and EfficientNet-B0 are more memory-efficient due to depthwise separable convolutions and MBConv blocks, but they introduce slightly more latency on our setup.

\par \noindent
% Target deployment on NVIDIA Jetson Orin Nano is feasible. The module features a 1024-core NVIDIA Ampere architecture GPU with 32 Tensor Cores, a 6-core Arm Cortex-A78AE CPU, and 8 GB RAM. Our model’s low compute cost and simple control flow make it suitable for CPU-only or TensorRT-accelerated FP16/INT8 deployment on the GPU. Given the 8 GB memory ceiling, we recommend quantization and modest pruning to keep peak memory comfortably below system limits, while MobileNetV2/EfficientNet variants can also be accelerated efficiently using Tensor Cores if a convolutional front-end is preferred.

\subsection{Hardware}
\par \noindent
All tests and measurements were performed on an NVIDIA Jetson Orin Nano. The module provides a 1024-core NVIDIA Ampere architecture GPU with 32 Tensor Cores, a 6-core Arm Cortex-A78AE CPU, and 8\,GB RAM. The reported latencies were obtained on this platform using batch size 1 and on-device acceleration.

\par \noindent
The models were trained on an Apple MPS and were exported to an ONNX format for inference on the Jeton Nano Orin

\subsection{Explainable AI}
\par \noindent
To better understand the decision making process of our model, we employed the Integrated Gradients method~\cite{sundararajan2017axiomatic} to attribute the model's predictions to specific pixels in the input image.
Integrated Gradients is a gradient based attribution technique that satisfies desirable axioms such as sensitivity and implementation invariance.
It computes feature attributions by integrating the gradients of the model's output with respect to the inputs, along a straight line path from a baseline to the actual input.
This allows us to highlight the regions of the input that most influenced the model's prediction, thereby providing interpretability for both correct and incorrect decisions.
\par \noindent
Mathematically, the attribution for the $i$-th input feature is computed as:
\begin{equation}
\nonumber
\mathrm{IntegratedGradients}_i^{}(x) = (x_i - x'_i) \times \frac{1}{m} \sum_{k=1}^m \frac{\partial F\left(x' + \frac{k}{m} (x - x')\right)}{\partial x_i}
\end{equation}
where $x$ is the input, $x'$ is the baseline input, and $F$ is the model's output function.
\par \noindent
This technique helped us identify which parts of the scene were most important for the model's localization decisions.
In our experiments, the attributions often concentrated on regions containing field lines and other spatial cues, suggesting that the model relies heavily on geometric features.
\end{multicols} % End columns for full-width figure
\begin{center}
    \includegraphics[scale=0.5]{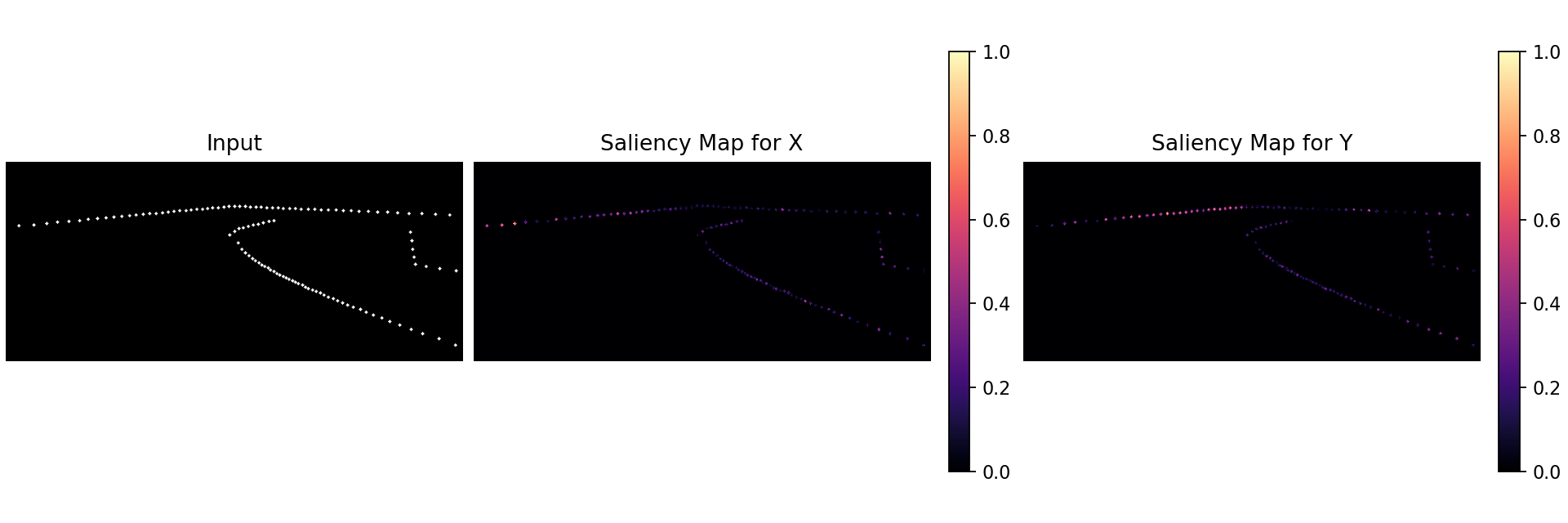}
    \captionof{figure}{Saliency Maps}\label{fig:comparison}
\end{center}
\begin{multicols}{2} % Resume columns after figure

\section{Failed Approaches}

\par \noindent
In our earlier methodology, we employed an odometry-based approach for the self-localization of the robots.
This system integrated wheel encoders and an Inertial Measurement Unit (IMU), complemented by a TF-Luna LiDAR sensor to estimate the depth between the robot and the basket.
However, the TF-Luna sensor demonstrated a limited effective range of approximately 3 to 4 meters and exhibited suboptimal accuracy within this range.
Additionally, a significant challenge arose due to the absence of a clearly distinguishable object or surface on the basket, which hindered reliable depth estimation.
\par \noindent
In the subsequent approach, we employed an object detection model based on YOLOv8n\cite{redmon2016yolo}, trained on a custom dataset to detect the basket.
This model was utilized not only for basket detection but also for correcting the robot’s yaw orientation to ensure it consistently faced the target.
To estimate the horizontal distance from the basket, we defined discrete zones, with each zone representing a specific distance range.
The corresponding zone, selected manually by the pilot based on visual assessment, was then used in conjunction with the detected orientation to determine the appropriate RPM for the shooting mechanism.
While this approach facilitated a simplified method for distance estimation, it was inherently dependent on manual selection and coarse approximations, which could lead to reduced accuracy and consistency.

\vspace{0.5cm}
{ \centering
  \includegraphics[scale=0.23]{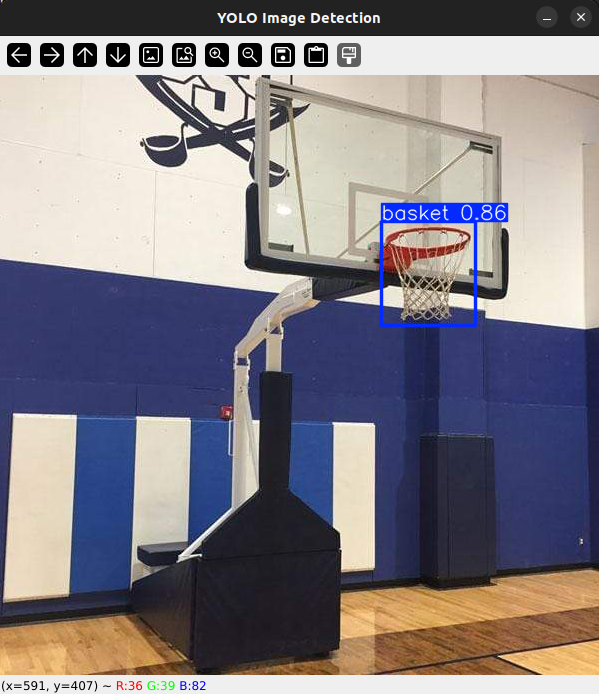}\\
  \captionof{figure}{basket detection}\label{fig:basket_detection}
}

\par \noindent
We improved upon the object detection approach by using the MiDaS\cite{ranftl2020midas} (Multiple Depth Estimation Accuracy with Single Network) deep learning model.
MiDaS is a designed for monocular depth estimation, i.e., predicting depth from a single image.
It uses a ResNet based encoder decoder architecture where the encoder extracts features from the input image and the decoder upsamples these features to produce a depth map.
The approach combines the YOLOv8 model with the MiDaS depth estimation framework for relative depth prediction.
The pipeline begins by loading an input image, followed by inference using the YOLO model to localize the basket and identify its center.
Simultaneously, MiDaS processes the same image to generate a dense relative depth map.
Once the basket is detected, the relative depth at the center of the bounding box is extracted from the depth map.
An interactive calibration step is used to relate MiDaS's relative depth to real world distance.
The user is prompted to enter known ground truth distances for three distinct positions.
These pairs of relative depth and real distance are then used to fit a linear regression model, enabling real-time conversion of relative depth to metric distance.
During tracking mode, the system periodically predicts the basket's distance using the learned regression model and overlays the result on the original image alongside the visualized depth map.
The output is displayed in a side by side format showing both RGB and color mapped depth views for easier interpretation.

\vspace{0.5cm}
{ \centering
  \includegraphics[width=0.4\textwidth]{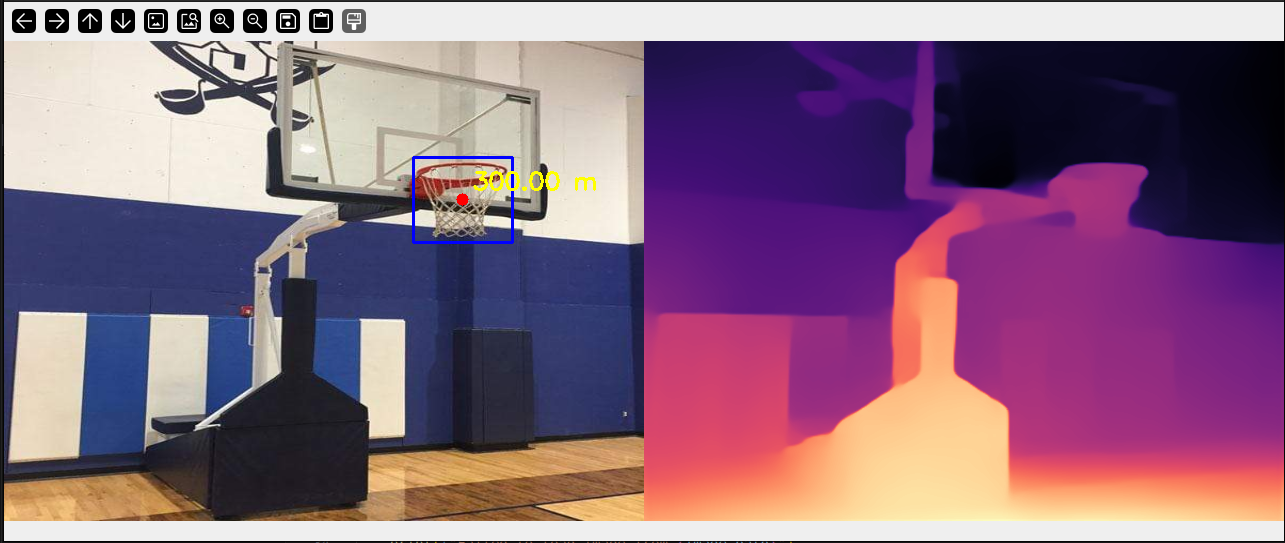}\\
  \captionof{figure}{basket depth estimation}\label{fig:basket_midas}
}

A key limitation of this approach was its reliance on prior calibration, which required manually measuring the distance between the robot and the basket at least three times before accurate predictions could be made.
In response to these constraints, we developed the current method presented in this paper, building upon insights gained from the previous techniques.
\section{Conclusion}  
\par \noindent
In this work, we presented a hybrid localization framework for autonomous basketball robots competing 
in the Robocon 2025 arena.
Our approach relies solely on floor images processed through a lightweight
feedforward neural network.
By preprocessing the images to highlight salient floor features and using a simple architecture, 
we achieved an average prediction error of approximately 0.06 meters, demonstrating the potential of 
learning-based localization methods.
This result shows that even with minimal sensor inputs and modest 
model complexity, it is possible to achieve sufficiently accurate localization.
Future improvements could 
focus on integrating other sensors such as IMU, wheel odometry, or LIDAR with the vision-based system 
through filtering techniques like Kalman or particle filters to achieve more robust and reliable 
localization.
Additionally, exploring more neural network architectures may help to improve prediction 
accuracy.
Finally, optimizing the model for deployment on embedded platforms with limited computational
resources through techniques such as pruning, quantization, or hardware acceleration could be worked upon.

\end{multicols}


\begin{thebibliography}{99}

\bibitem{gazebo}
N. Koenig and A. Howard, "Design and use paradigms for Gazebo, an open-source 
multi-robot simulator," 2004 IEEE/RSJ International Conference on Intelligent Robots 
and Systems (IROS) (IEEE Cat. No.04CH37566), Sendai, Japan, 2004, pp. 2149-2154 vol.3, 
doi: 10.1109/IROS.2004.1389727.
\bibitem{ismail2012}
M. Ismail \textit{et al.}, ``Soccer robot localization based on sensor fusion from odometry and omnivision,'' 2012.

\bibitem{kingma2014adam}
Diederik P. Kingma and Jimmy Ba.
\textit{Adam: A method for stochastic optimization}. 
arXiv preprint arXiv:1412.6980, 2014.

\bibitem{liu2009}
Q. Liu \textit{et al.}, ``A self-localization method through pose point matching for autonomous soccer robot based on omni-vision,'' in \textit{Proceedings of RoboCup Symposium}, 2009.

\bibitem{lu2013}
H. Lu \textit{et al.}, ``Robust and real-time self-localization based on omnidirectional vision for soccer robots,'' 2013.

\bibitem{luo2020}
X. Luo \textit{et al.}, ``Robot detection and localization based on deep learning,'' 2020.

\bibitem{macenski2022ros2}
Steve Macenski, Tully Foote, Brian Gerkey, Chris Lalancette, and William Woodall.
\textit{Robot Operating System 2: Design, architecture, and uses in the wild}. 
Science Robotics, vol. 7, no. 66, 2022. 

\bibitem{nadiri2025}
A. Nadiri \textit{et al.}, ``IMU-stabilized bird’s-eye view localization for humanoid soccer robots,'' 2025.

\bibitem{ribeiro2016}
M. Ribeiro \textit{et al.}, ``Fast computational processing for mobile robots self-localization,'' 2016.

\bibitem{tian2010}
Y. Tian \textit{et al.}, ``Self-localization of humanoid robots with fish-eye lens in a soccer field,'' 2010.

\bibitem{watanabe2020}
Y. Watanabe \textit{et al.}, ``Model-based self-localization with genetic algorithm for omnidirectional vision,'' 2020.

\bibitem{sundararajan2017axiomatic}
M.~Sundararajan, A.~Taly, and Q.~Yan, ``Axiomatic attribution for deep networks,'' 
in \emph{Proceedings of the 34th International Conference on Machine Learning (ICML)}, 
vol.~70, pp.~3319--3328, 2017.

\bibitem{tola2024understanding}
D. Tola and P. Corke, ``Understanding URDF: A dataset and analysis,'' \textit{IEEE Robotics and Automation Letters}, vol.~9, no.~5, pp.~4479--4486, 2024.

\bibitem{redmon2016yolo}
J.~Redmon, S.~Divvala, R.~Girshick, and A.~Farhadi, 
``You only look once: Unified, real-time object detection,'' 
in \textit{Proceedings of the IEEE Conference on Computer Vision and Pattern Recognition (CVPR)}, 
2016, pp.~779--788.
\bibitem{ranftl2020midas}
R.~Ranftl, A.~Boedt, and V.~Koltun, 
``Towards robust monocular depth estimation: Mixing datasets for zero-shot cross-dataset transfer,'' 
\textit{IEEE Transactions on Pattern Analysis and Machine Intelligence}, 
vol.~44, no.~3, pp.~1623--1637, 2020.
\bibitem{sandler2018mobilenetv2}
M.~Sandler, A.~Howard, M.~Zhu, A.~Zhmoginov, and L.-C.~Chen,
``MobileNetV2: Inverted residuals and linear bottlenecks,''
in \textit{Proceedings of the IEEE Conference on Computer Vision and Pattern Recognition (CVPR)},
2018, pp.~4510--4520.
\bibitem{tan2019efficientnet}
M.~Tan and Q.~V.~Le,
``EfficientNet: Rethinking model scaling for convolutional neural networks,''
in \textit{Proceedings of the 36th International Conference on Machine Learning (ICML)},
PMLR, 2019, pp.~6105--6114.

\end{thebibliography}
\end{document}